\documentclass[11pt]{article}
\usepackage[totalwidth=465pt,totalheight=645pt]{geometry}
\usepackage{graphicx,amsmath,amssymb,bbm,algorithmic,algorithm}
\newcommand{\1}{{\mathbbm{1}}}

\title{Efficient Learning of Sparse Conditional Random Fields for Supervised Sequence Labelling}
\author{Nataliya Sokolovska, Thomas Lavergne, Olivier Capp\'e and Fran\c cois Yvon\thanks{N. Sokolovska and O. Capp\'e are with Telecom ParisTech and LTCI CNRS, 46 rue Barrault, 75013 Paris, France; T. Lavergne and F. Yvon are with Universit\'e Paris-Sud 11 and LIMSI CNRS, 91403 Orsay, France.}}
\date{}

\begin{document}

\maketitle

\begin{abstract}
  Conditional Random Fields (CRFs) constitute a popular and efficient approach for supervised
  sequence labelling. CRFs can cope with large description spaces and can integrate some form of
  structural dependency between labels.
  %
  In this contribution, we address the issue of efficient feature selection for CRFs based on
  imposing sparsity through an ${\ell^1}$ penalty. We first show how sparsity of the parameter set can be
  exploited to significantly speed up training and labelling. We then introduce coordinate descent
  parameter update schemes for CRFs with ${\ell^1}$ regularization. We finally provide some empirical
  comparisons of the proposed approach with state-of-the-art CRF training strategies. In
  particular, it is shown that the proposed approach is able to take profit of the sparsity to
  speed up processing and handle larger dimensional models.
\end{abstract}

\section{Introduction}
\label{sec:introduction}
Conditional Random Fields (CRFs), originally introduced in
\cite{Lafferty01conditional}, constitute a popular and effective approach for
supervised structure learning tasks involving the mapping between complex
objects such as strings and trees. 
An important property of CRFs is their ability to cope with large and
redundant feature sets and to integrate some form of structural dependency
between output labels. Directly modeling the conditional probability
of the label sequence given the observation stream allows to
explicitly integrate complex dependencies that can not directly be accounted
for in generative models such as Hidden Markov Models (HMMs). Results
presented in section~\ref{ssec:nettalk} will illustrate this ability
to use large sets of redundant and non-causal features. 

Training a CRF amounts to solving a convex optimization problem: the
maximization of the penalized conditional log-likelihood function. 
For lack of an analytical solution however, the CRF training task requires
numerical optimization and implies to 
repeatedly perform inference over the entire training set during the computation
of the gradient of the objective function.

This is where the modeling of structure takes its toll: for general dependencies,
exact inference is intractable and approximations have to be considered. In the
simpler case of linear-chain CRFs, modeling the interaction between pairs of
adjacent labels makes the complexity of inference grow quadratically with the
size of label set: even in this restricted setting, training a CRF remains a
computational burden, especially when the number of output labels is large.

Introducing structure has another, less studied, impact on the number of
potential features that can be considered. It is possible, in a linear-chain
CRF, to introduce features that simultaneously test the values of adjacent
labels and some property of the observation. In fact, these features often
contain valuable information \cite{Peng06information}. However, their number
scales quadratically with the number of labels, yielding both a computational
(feature functions have to be computed, parameter vectors have to be stored in
memory) and an estimation problem.

The estimation problem stems from the need to estimate large parameter vectors
based on sparse training data. Penalizing the objective function with the
$\ell^2$ norm of the parameter vector is an effective remedy to overfitting;
yet, it does not decrease the number of feature computations that are needed. In
this paper, we consider the use of an alternative penalty function, the $\ell^1$
norm, which yields much sparser parameter vectors \cite{Tibshirani96Lasso}\footnote{To be more precise, we consider in the following a mixed penalty which involves both $\ell^1$ and squared $\ell^2$ terms, also called the \emph{elastic net penalty} \cite{Zhou05elastic}. The sparsity of the solution is however controlled mostly by the amount of $\ell^1$ regularization.}. As
we will show, inducing a sparse vector not only reduces the number of feature
functions that need to be computed, but it can also reduce the time needed to perform
parameter estimation and decoding.

The main shortcoming of the $\ell^1$ regularizer is that the objective function
is no longer differentiable everywhere, challenging the use of gradient-based
optimization algorithms. Proposals have been made to overcome this difficulty:
for instance, the orthant-wise limited-memory quasi-Newton algorithm \cite{Andrew07L1Reg}
uses the fact that the $\ell^1$ norm remains differentiable when
restricted to regions in which the sign of each coordinate is fixed (an ``orthant''). Using
this technique, \cite{Gao07comparativeNLP} reports test performance that are on
par with those obtained with a $\ell^2$ penalty, albeit with more compact
models. Our first contribution is to show that even in this situation (equivalent test
performance), the  $\ell^1$ regularization may be preferable as sparsity in the parameter
set can be exploited to reduce the computational cost associated with parameter training
and label inference.

For parameter estimation, we consider an alternative optimization approach, which generalizes to
CRFs the proposal of \cite{Friedman07coordinatedesc} (see also \cite{Dudik04performance,Krishnapuram05sparse}).  In a
nutshell, optimization is performed in a coordinate-wise fashion, based on an analytic solution to
the unidimensional optimization problem. In order to tackle realistic problems, we propose an efficient blocking
scheme in which the coordinate-wise updates are applied simultaneously to a properly selected group
(or block) of parameters. Our main methodological contributions are thus twofold: (i) a fast
implementation of the training and decoding algorithms that uses the sparsity of parameter vectors
and (ii) a novel optimization algorithm for using $\ell^1$ penalty with CRFs. These two ideas
combined together offer the opportunity of using very large ``virtual'' feature sets for which only
a very small number of features are effectively selected. As will be seen (in Section~\ref{sec:sparsity}), this situation is frequent in typical natural language processing applications, particularly when the number of possible labels is large. Finally, the proposed algorithm
has been implemented as C code and validated through experiments on artificial and
real-world data. In particular, we provide detailed comparisons, in terms of numerical
efficiency, with solutions traditionally used for $\ell^2$ and $\ell^1$ penalized training of CRFs
in publicly available software such as CRF++ \cite{kudo05crfpp}, 
CRFsuite \cite{okazaki07crfsuite} and crfsgd \cite{bottou07sgd}.

The rest of this paper is organized as follows. In Section \ref{sec:sparsity},
we introduce our notations and restate more precisely the issues we wish to
address, based on the example of a simple natural language processing
task. Section~\ref{sec:inference} discusses the algorithmic gains that are
achievable when working with sparse parameter vectors. We then study, in Section
\ref{sec:estimation}, the training algorithm used to achieve sparsity, which
implements a coordinate-wise descent procedure. Section~\ref{sec:discus} discusses
our contributions with respect to related work. And finally, Section \ref{sec:exper} presents
our experimental results, obtained both on simulated data, a
phonetization task, and a named entity recognition problem.

\section{Conditional Random Fields and Sparsity \label{sec:sparsity}}
Conditional Random Fields \cite{Lafferty01conditional,Sutton06introduction} are based on the following discriminative probabilistic model 
\begin{equation}
p_\theta(\mathbf{y}|\mathbf{x}) = \frac{1}{Z_\theta(\mathbf{x})} \exp \Bigg\{\sum_{t=1}^T \sum_{k=1}^K \theta_k  f_k(y_{t-1}, y_t, x_t) 
\Bigg\}
\label{eq:crf}
\end{equation}
where $\mathbf{x} = (x_1,\dots, x_T)$ denotes the input sequence and $\mathbf{y} = (y_1,\dots,
y_T)$ is the output sequence, also referred to as the sequence of labels. $\{f_k\}_{1\leq k
  \leq K}$ is an arbitrary set of feature functions and $\{\theta_k\}_{1\leq k \leq K}$ are the
associated real-valued parameter values\footnote{Note that various conventions are found in the
  literature regarding the treatment of the initial term (with index $t=1$) in~\eqref{eq:crf}. Many
  authors simply ignore the term corresponding to the initial position $t=1$ for so-called (see
  below) bigram features. In our implementation, $y_{0}$ refers to a particular (always observed)
  label that indicates the beginning of the sequence. In effect, this adds a few parameters that are
  specific to this initial position. However, as the impact on performance is usually
  negligible, we omit this specificity in the following for the sake of simplicity.}. The CRF form
considered in~(\ref{eq:crf}) is sometimes referred to as linear-chain CRF, although we stress that
it is more general, as $y_t$ and $x_t$ could be composed not directly of the individual sequence
tokens, but on sub-sequences (e.g., trigrams) or other localized characteristics. We will denote by
$Y$, $X$, respectively, the sets in which $y_t$ and $x_t$ take their values. The normalization
factor in~\eqref{eq:crf} is defined by
\begin{equation}
Z_\theta(\mathbf{x}) = \sum_{\mathbf{y} \in Y^T} \exp \Bigg\{\sum_{t=1}^T\sum_{k=1}^K \theta_k f_k(y_{t-1}, y_t, x_t) \Bigg\}.
\end{equation}

The most common choice of feature functions is to use binary tests such that $f_k(y_{t-1}, y_t,
x_t)$ is one only when the triplet $(y_{t-1}, y_t, x_t)$ is in a
particular configuration. In this setting, the number of parameters $K$ is equal to
$|Y|^2 \times |X|_{\text{train}}$, where $|\cdot|$ denotes the cardinal and $|X|_{\text{train}}$
refers to the number of configurations of $x_t$ observed in the training
set. As discussed in Section~\ref{sec:inference} below, the bottleneck when performing inference
is the computation of the pairwise conditional probabilities
$p_\theta(y_{t-1}=y,y_{t}=y'|\mathbf{x})$, for $t=1,\dots,T$ and $(y,y')\in Y^2$ for all training
sequences, which involves a number of operations that scales as $|Y|^2$ times the number of
training tokens. Thus, even in moderate size applications, the number of parameters can be very
large and the price to pay for the introduction of sequential dependencies in the model is rather high, explaining why it is hard to train CRFs with dependencies between more
than two adjacent labels.

To motivate our contribution, we consider below a 
moderate-size natural language processing task, namely a word phonetization task
based on the Nettalk dictionary \cite{Sejnowski87}, where $|Y|$ (the number of
phonemes) equals 53 and $|X|$ is 26 (one value for each English letter).
For this task, we use a CRF that involves two types of features
functions, which we refer to as, respectively, \emph{unigram features}, $\mu_{y,x}$, and \emph{bigram features}, $\lambda_{y',y,x}$. These are such that
\begin{multline}
   \sum_{k=1}^K \theta_k  f_k(y_{t-1}, y_t, x_t) = \sum_{y \in Y, x \in X} \mu_{y,x} \mathbf{1}(y_t = y, x_t = x) \\
   + \sum_{(y',y) \in Y^2, x \in X} \lambda_{y',y,x} \mathbf{1}(y_{t-1} = y', y_t = y, x_t = x)
   \label{eq:mu_lambda} 
\end{multline}
where $\mathbf{1}(\mathrm{cond.})$ is equal to 1 when the condition is verified and to 0 otherwise.

The use of the sole unigram features $\{\mu_{y,x}\}_{(y,x) \in Y \times X}$ would result in a model
equivalent to a simple bag-of-tokens position-by-position logistic regression model. On the other
hand, bigram features $\{\lambda_{y',y,x}\}_{(y,x) \in Y^2 \times X}$ are helpful in modelling
dependencies between successive labels. The motivations for using simultaneously both types of
feature functions and the details of this experiment are discussed in Section~\ref{sec:exper}.  In
the following, by analogy with the domain of constrained optimization, we refer to the subset of
feature functions whose multiplier is non-zero as the ``active'' features.


\begin{figure}[hbt]
\centering
\includegraphics[width=0.35\textwidth]{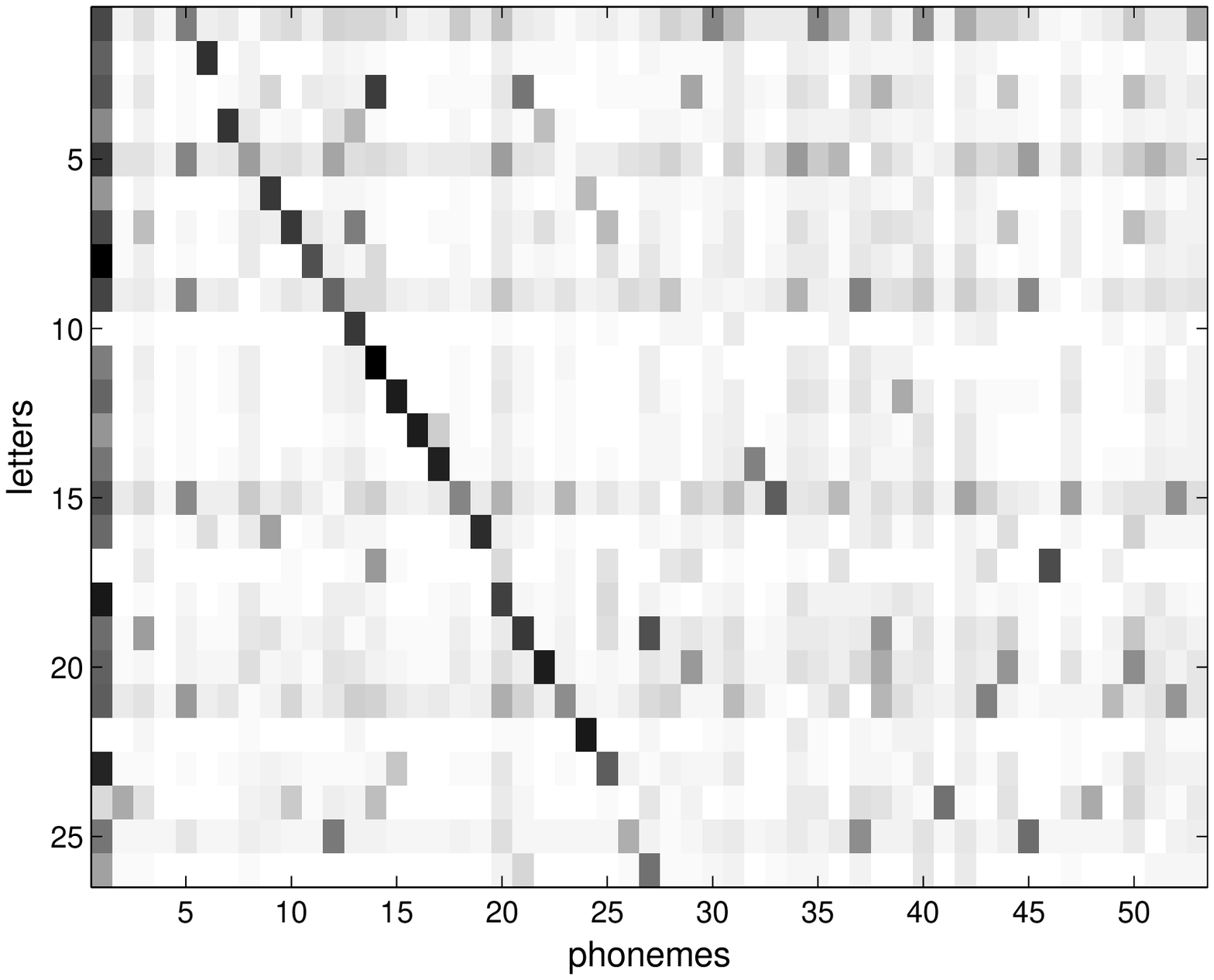} \qquad \includegraphics[width=0.35\textwidth]{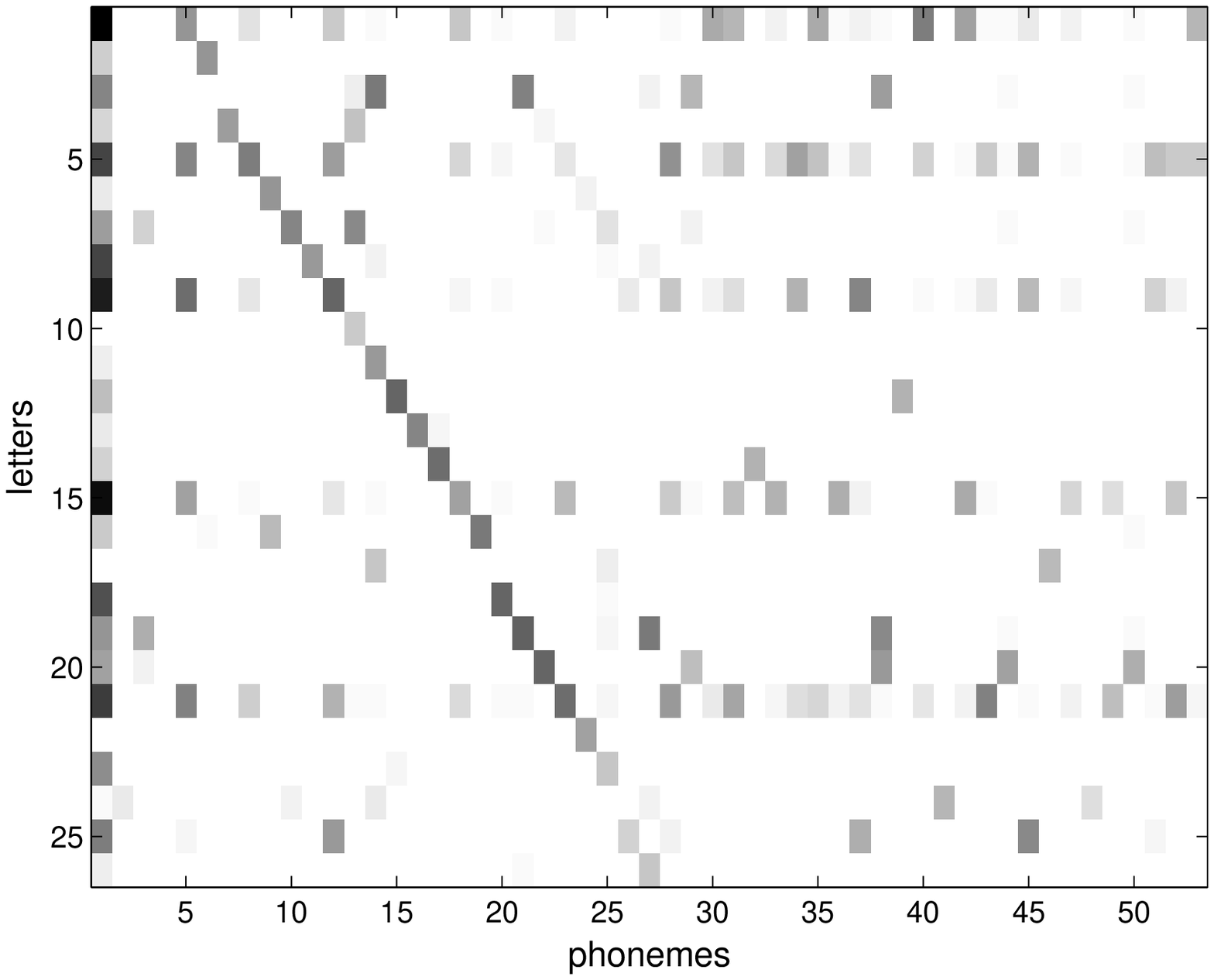}
\caption{The $\ell^1$ norm of the parameters estimated with standard $\ell^2$-regularized maximum likelihood for the Nettalk task. Above: $|\mu_{y,x}|$ for the  53 phonemes $y$ and 26 letters $x$. Below: $\sum_{y'} |\lambda_{y',y,x}|$ for the 53 phonemes $y$ and 26 letters $x$.}
\label{fig:crfl2_nettalk} 
\end{figure}

Figure~\ref{fig:crfl2_nettalk} allows to visualize the magnitude of the parameter
vectors obtained with the $\ell^2$-regularized maximum likelihood
approach. Sparsity is especially striking in the case of the $\lambda_{y',y,x}$
parameters which are, by far, the most numerous ($53^2 \times 26$). Another
observation is that this sparsity pattern is quite correlated with the
corresponding value of $|\mu_{y,x}|$: in other words, most sequential
dependencies $\lambda_{y',y,x}$ are only significant when the associated
marginal factor $\mu_{y,x}$ is. This suggests to take a closer look at the
internal structure of the feature set.

From this picture, one would expect to attain the same classification accuracy
with a much reduced set of feature functions using an appropriate feature
selection approach. These preliminary considerations motivate some of the
questions that we try to answer in this contribution: (1) Is it possible to take
profit of the fact that a large proportion of the parameters are null to speed up computations? (2)
How can we select features in a principled way during the parameter estimation
phase? (3) Can sparse solutions also result in competitive test accuracy?

\section{Fast Computations in Sparse CRFs \label{sec:inference}}
\subsection{Computation of the Objective Function and its Gradient}
Given $N$ independent labelled sequences $\{\mathbf{x}^{(i)},
\mathbf{y}^{(i)}\}_{i=1}^N$, where both $\mathbf{x}^{(i)}$  and
$\mathbf{y}^{(i)}$ contain $T^{(i)}$ symbols, conditional maximum likelihood estimation is based on the minimization, with respect to $\theta$, of
\begin{align}
l(\theta) &= - \sum_{i=1}^N \log p_\theta(\mathbf{y}^{(i)}|\mathbf{x}^{(i)}) \nonumber \\
&= \sum_{i=1}^N \left\{\log Z_\theta(\mathbf{x}^{(i)}) - \sum_{t=1}^{T^{(i)}} \sum_{k=1}^K \theta_k f_k(y_{t-1}^{(i)}, y_{t}^{(i)}, x_t^{(i)}) \right\}
\label{eq:logloss}
\end{align}
The function $l(\theta)$ is recognized as the negated conditional log-likelihood of the observations and will be referred to in the following as the \emph{logarithmic loss function}. 
This term is usually complemented with an additional regularization term so as
to avoid overfitting (see Section~\ref{sec:estimation} below). The gradient of $l(\theta)$ is given by
\begin{align}
\label{eq:1rstorder}
\frac{\partial l( \theta)}{\partial \theta_k} = & \sum_{i=1}^N \sum_{t=1}^{T^{(i)}} \operatorname{E}_{p_\theta(\mathbf{y} | \mathbf{x}^{(i)})} f_k(y_{t-1}, y_t, x_t^{(i)})\nonumber \\ & - \sum_{i=1}^N \sum_{t=1}^{T^{(i)}} f_k(y_{t-1}^{(i)}, y_t^{(i)}, x_t^{(i)})
\end{align}
where $\operatorname{E}_{p_\theta(\mathbf{y} | \mathbf{x}^{(i)})}$ denotes the conditional expectation given the observation sequence, i.e.
\begin{multline}
  \operatorname{E}_{p_\theta(\mathbf{y} | \mathbf{x}^{(i)})} f_k(y_{t-1}, y_t, x_t^{(i)}) = \\
  \sum_{(y',y) \in \mathcal{Y}^2} f_k(y, y', x_t^{(i)}) \operatorname{P}_\theta(y_{t-1} = y', y_t = y | \mathbf{x}^{(i)})
\label{eq:condexp}
\end{multline}
Although $l(\theta)$ is a smooth convex function, it has to be
optimized numerically. The computation of its gradient implies to
repeatedly compute the conditional expectation in~\eqref{eq:condexp} for all
input sequences $\mathbf{x}^{(i)}$ and all positions $t$.

\subsection{Sparse Forward-Backward Algorithm}
\label{sec:fb}
The standard approach for computing the conditional probabilities in CRFs is inspired by the
forward-backward algorithm for hidden Markov models: in the case of the parameterization
of~\eqref{eq:mu_lambda}, the algorithm implies the computation of the forward
\begin{equation}
\begin{cases}
\alpha_1(y) = \exp(\mu_{y, x_1} + \lambda_{y_0,y,x_1}) \\
\alpha_{t+1}(y) = \sum_{y'}\alpha_t(y') \exp(\mu_{y, x_{t+1}} +  \lambda_{y',y,x_{t+1}})
\end{cases} 
\label{eq:forw}
\end{equation}
and the backward recursion
\begin{equation}
\begin{cases}
\beta_{T_i}(y) = 1\\
\beta_t(y') = \sum_{y}\beta_{t+1}(y) \exp(\mu_{y, x_{t+1}} +  \lambda_{y',y,x_{t+1}}),
\end{cases} 
\label{eq:backw}
\end{equation}
for all indices $1 \leq t \leq T$ and all labels $y \in Y$. Then, $Z_\theta(\mathbf{x}) = \sum_{y} \alpha_{T}(y)$ and the pairwise probabilities $\operatorname{P}_\theta(y_{t} = y', y_{t+1} = y | \mathbf{x})$ are given by
\[
   \alpha_{t}(y')\exp(\mu_{y, x_{t+1}} + \lambda_{y',y,x_{t+1}}) \beta_{t+1}(y) / Z_\theta(\mathbf{x})
\]
These recursions require a number of operations that grows quadratically with $|Y|$.

Let us now consider the case where the set of bigram features $\{\lambda_{y',y,x_{t+1}}\}_{(y',y)\in Y^2}$ is sparse with only $r(x_{t+1}) \ll |Y|^2$ non null values and define the matrix
$$
  M_{t+1}(y',y) = \exp (\lambda_{y',y,x_{t+1}}) - 1
$$
Observe that $M_{t+1}(y',y)$ also is a sparse matrix and that the forward and backward equations
may be rewritten as
\begin{align}
  & \alpha_{t+1}(y) = \exp(\mu_{y, x_{t+1}}) \bigg\{ \sum_{y'}\alpha_t(y') + \sum_{y'}\alpha_t(y') M_{t+1}(y',y) \bigg\} \nonumber \\
  & \beta_t(y') =  \sum_{y} v_{t+1}(y) + \sum_{y} M_{t+1}(y',y) v_{t+1}(y)
  \label{eq:sparsefb}
\end{align}
where $v_{t+1}(y) = \beta_{t+1}(y) \exp(\mu_{y, x_{t+1}})$. The resulting computational savings stem from the fact that the vector matrix
products in~\eqref{eq:sparsefb} now only involve the sparse matrix $M_{t+1}(y',y)$. This means that they can be
computed, using an appropriate sparse matrix implementation, with exactly $r(x_{t+1})$
multiplications instead of $|Y|^2$. If the set $\{\mu_{y,x_{t+1}}\}_{y\in Y}$ of unigram features is also sparse, one
may use a similar idea although the computation savings will in general be less significant.

Using the implementation outlined in~\eqref{eq:sparsefb}, the complexity of the forward-backward
procedure for the sequence $\mathbf{x}^{(i)}$ can be reduced from $T^{(i)} \times |Y|^2$ to the
cumulated sizes of the feature sets encountered at each position along the sequence. Thus,
the complexity of the forward-backward procedure is proportional to the average number
of active features per position in the parameter set rather than to the actual number of
potentially active features. This observation suggests that it might even be possible to use some
longer term dependencies between labels, as long as only a few of them are active simultaneously.

It should be stressed that both CRF++ \cite{kudo05crfpp} and crfsgd \cite{bottou07sgd} use
logarithmic computation in the forward-backward recursions, that is, updating $\log \alpha_{t}(y)$
and $\log \beta_{t}(y)$ rather than $\alpha_{t}(y)$ and $\beta_{t}(y)$ in~\eqref{eq:forw}
and~\eqref{eq:backw}. The advantage of logarithmic computations is that numerical over/underflows
are avoided whatever the length $T^{(i)}$ of the sequence, whereas the linear form of~\eqref{eq:forw}
and~\eqref{eq:backw} is only suitable for sequences whose length is less than a few tens. On the
other hand, logarithmic computation is not the only way of avoiding numerical issues (the
``scaling'' solution traditionally used for HMMs \cite{cappe-moulines-ryden-2005} applies as well
here) and is very inefficient from an implementation point of view due to the repeated calls to the
\verb+exp+ function (see Section \ref{ssec:nettalk}). This being said, when logarithmic
computations are used, \eqref{eq:sparsefb} may be used in a similar fashion to reduce the
complexity of the logarithmic update when $M_{t+1}(y',y)$ is sufficiently sparse.

Note that although we focus in this paper on the complexity of the training phase, the above idea
may also be used to reduce the computational burden associated with Viterbi (or
optimal sequence-wise) decoding. Indeed, at position $t+1$, the forward pass in the Viterbi recursion amounts to computing
\begin{equation}
\epsilon_{t+1}(y) = \max_{y'\in Y}\left\{ \epsilon_t(y') + \lambda_{y',y,x_{t+1}} \right\} + \mu_{y,x_{t+1}}
\label{eq:viterb}
\end{equation}
where $\epsilon_t(y)$ denotes the conditional log-likelihood of the optimal 
labelling of the $t$ first tokens subject to the constraint that the last label is $y$ (omitting
the constant $-\log Z_\theta(\mathbf{x})$ which is common to all possible
labellings). Assuming that $A(y,x_{t+1}) = \{y'\in Y: \lambda_{y',y,x_{t+1}} \neq 0\}$ is limited to
a few labels, it is possible to implement~\eqref{eq:viterb} as
\begin{equation*}
\epsilon_{t+1}(y) = \max \bigg\{ \max_{y' \in \bar{A}(y,x_{t+1})} \epsilon_t(y'),
  \max_{y'\in A(y,x_{t+1})}\left( \epsilon_t(y') + \lambda_{y',y,x_{t+1}} \right) \bigg\} + \mu_{y,x_{t+1}}
\end{equation*}
where $\bar{A}(y,x_{t+1})$ denotes the elements of $Y$ that are not in the current active set $A(y,x_{t+1})$ of bigram features. Hence the number of required additions now is of the order of the number of active features, $|A(y,x_{t+1})|$, rather than equal to the number of labels $|Y|$.


\section{Parameter Estimation Using Blockwise Coordinate Descent}
\label{sec:estimation}

\subsection{Regularization}
The standard approach for parameter estimation in CRFs consists in minimizing
the logarithmic loss $l(\theta)$ defined by~\eqref{eq:logloss} with an
additional squared $\ell^2$ penalty term $\frac{\rho_2}{2} \|\theta
\|_2^2$, where $\rho_2$ is a regularization parameter. The objective function is
then a smooth convex function to be minimized over an unconstrained parameter
space. Hence, any numerical optimization strategy may be used for this purpose
and popular solutions include limited memory BFGS (L-BFGS) \cite{Nocedal89lbfgs} or conjugate
gradient. Note that it is important however, to avoid numerical optimizers that require the full Hessian matrix (e.g., Newton's algorithm) or approximations of it due to the size of the parameter vector in usual applications of CRFs.


In the following, we consider the elastic net penalty \cite{Zhou05elastic} which combines $\ell^1$ and $\ell^2$ regularizers
and yields an objective function 
\begin{equation} 
  l(\theta) + \rho_1 \|\theta \|_1 + \frac{\rho_2}{2} \|\theta \|_2^2
\label{eq:elasticnet}
\end{equation}
where $\rho_1$ and $\rho_2$ are regularization parameters. The use of both types of penalty terms seems preferable in log-linear conditional models, as it makes it possible to control both the number of non zero coefficients (through $\rho_1$) and to avoid the numerical problems that might occur in large dimensional parameter settings if the magnitude of the $\theta_k$s is not sufficiently constrained by the penalty.

It may also be rewarding to look for some additional information in hierarchical and group
structure of the data. An example is the group lasso estimator, introduced in
\cite{Meier08grouplasso} as an extension of the lasso. The motivation for the group lasso is to
select not individual variables, but whole blocks of variables. Typically, the penalty takes the
form of the sum of the $\ell^2$ norms of predefined blocks of the parameter vector. This idea has
been further extended in \cite{Zhao09compositepenalty} under the name of Composite Absolute
Penalties (CAP) for dealing with more complex a priori parameter hierarchies while still retaining
an overall convex penalty term. Although our approach could be suitable for this more complex
choices of the penalty function, we restrict ourselves in the following to the case of the elastic
net penalty.

\subsection{Coordinate Descent}
The objective function in~\eqref{eq:elasticnet} is still convex but not
differentiable everywhere due to the $\ell^1$ penalty term. Although different
algorithms have been proposed to optimize such a criterion, we believe that the
coordinate-wise approach of \cite{Friedman08GLM} has a strong potential for CRFs
as the update of the parameter $\theta_k$ only involves carrying out the forward-backward recursions for those sequences that contain symbols $x$ such that at
least one of the values $\{f_k(y',y,x)\}_{(y,y')\in Y^2}$ is non null, which is
most often much smaller than the total number of training sequences. This
algorithm operates by first considering a local quadratic approximation of the
objective function around the current value $\bar{\theta}$:
\begin{align}
& l_{k,\bar{\theta}}( \theta_k) = C^{st} + \frac{\partial l( \bar{\theta})}{\partial \theta_k}(\theta_k - \bar{\theta}_k) + \frac{1}{2} \frac{\partial^2 l( \bar{\theta})}{\partial \theta_k^2}(\theta_k - \bar{\theta}_k)^2 \nonumber \\
& \qquad \qquad \qquad + \rho_1 |\theta_k| + \frac{\rho_2}{2} \theta_k^2
\label{eq:approx}
\end{align}
Then, the minimizer of the approximation~\eqref{eq:approx} is easily found to be
\begin{equation}
  \theta_k = \frac{s\left\{\frac{\partial^2 l( \bar{\theta})}{\partial \theta_k^2} \bar{\theta}_k - \frac{\partial l( \bar{\theta})}{\partial \theta_k}, \rho_1\right\}}{\frac{\partial^2 l( \bar{\theta})}{\partial \theta_k^2} + \rho_2}
\label{eq:coord_update}
\end{equation}
where $s$ is the soft-thresholding function defined by
\begin{equation}
  s(z,\rho) = \begin{cases} z - \rho & \text{if $z > \rho$} \\
  z + \rho & \text{if $z < -\rho$} \\
  0 & \text{otherwise}
  \end{cases}
\label{eq:coord_update:st}
\end{equation}

Interestingly, \cite{Dudik04performance} originally proposed a similar idea but based on a
different local approximation of the behavior of the logarithmic loss. In
\cite{Dudik04performance}, the local behavior of the function $l(\theta)$ is approximated under a
form that is equivalent to the first-order only and leads to a closed-form coordinate-wise
optimization formula as well. This approximation is however explicitly based on the fact that each
parameter $\theta_k$ is multiplied by a function that takes its values in $\{0, 1\}$. This property
is not verified for CRF, since $\theta_k$ is multiplied by $\sum_{t=1}^T f_k(y_{t-1}, y_t, x_t)$,
which can be more than $1$ if the corresponding feature is observed at several positions in the
sequence.

\subsection{Coordinate Descent for CRFs}
To apply the algorithm described above for CRFs, one needs to be able to
compute~\eqref{eq:coord_update:st}, which requires to evaluate the first and second
order derivatives of $l(\theta)$. If the first order derivative is readily
computable using the forward-backward recursions described in
Section~\ref{sec:fb} and~\eqref{eq:1rstorder}, the exact computation of the
second derivative is harder for CRFs.
In fact, standard computations show that the diagonal term of the Hessian is
\begin{multline}
\frac{\partial^2 l(\theta)}{\partial \theta_k^2} =  \sum_{i=1}^N \Bigg\{ E_{p_\theta(\mathbf{y} | \mathbf{x}^{(i)})} \left( \sum_{t=1}^{T^{(i)}} f_k(y_{t-1}, y_t, x_t^{(i)}) \right)^2 \\
  - \left( E_{p_\theta(\mathbf{y} | \mathbf{x}^{(i)})} \sum_{t=1}^{T^{(i)}} f_k(y_{t-1}, y_t, x_t^{(i)}) \right)^2  \Bigg\}
  \label{eq:2ndorder}
\end{multline}
The first term is problematic as it involves the conditional expectation of a
square which cannot be computed only from the pairwise probabilities
$\operatorname{P}_\theta(y_{t-1}=y',y_t=y | \mathbf{x}^{(i)})$ returned by the
forward-backward procedure. It can be shown (see Chapter 4 of
\cite{cappe-moulines-ryden-2005} and \cite{cappe-moulines-2005a})
that~\eqref{eq:2ndorder} can be computed using auxiliary recursions related to
the usual forward recursion with an overall complexity of order $|Y|^2 \times
T^{(i)}$ per sequence. Unfortunately, this recursion is specific for each index $k$
and cannot be shared between parameters. As will be shown below, sharing (part of) the computations between parameters is
desirable feature for handling non trivial CRFs; we thus propose to use
instead the approximation
\begin{align}
\frac{\partial^2 l(\theta)}{\partial \theta_k^2} \approx & 
 \sum_{i=1}^N \sum_{t=1}^{T^{(i)}} \operatorname{E}_{p_\theta(\mathbf{y} | \mathbf{x}^{(i)})}f_k^2(y_{t-1}, y_t, x_t^{(i)}) \nonumber \\ & - \left( \operatorname{E}_{p_\theta(\mathbf{y} | \mathbf{x}^{(i)})}f_k(y_{t-1}, y_t, x_t^{(i)}) \right)^2   
\label{eq:2ndorder_approx}
\end{align}
This approximation amounts to assuming that, \emph{given} $\mathbf{x}^{(i)}$,
$f_k(y_{t-1}, y_t, x_t^{(i)})$ and $f_k(y_{s-1}, y_s, x_s^{(i)})$ are
uncorrelated when $s \neq t$. Note that this approximation is exact when the
feature $f_k$ is only active at one position along the sequence. It is likely
that the accuracy of this approximation is reduced when $f_k$ is active twice,
especially if the corresponding positions $s$ and $t$ are close.

The proposed coordinate descent algorithm applied to CRFs is
summarized as Algorithm~\ref{alg:crfcoord}.


\begin{algorithm}
\caption{Coordinate Descent for CRF}
\label{alg:crfcoord}
\begin{algorithmic}
\WHILE {Convergence criterion is not met}
\FOR {$k = 1:K$}
\FOR {Sequences for which $f_k$ is active}
\STATE Perform sparse forward-backward.
\ENDFOR
\STATE Compute $\partial l (\theta)/\partial \theta_k$ and $\partial^2 l(\theta)/\partial \theta_k^2$ from~(\ref{eq:1rstorder})--\eqref{eq:2ndorder_approx}.
\STATE Update $\theta$ according to~\eqref{eq:coord_update}--\eqref{eq:coord_update:st}.
\ENDFOR
\ENDWHILE
\end{algorithmic}
\end{algorithm}


A potential issue with this algorithm is the fact that, in contrast to the
logistic regression case considered in~\cite{Friedman08GLM}, we are using an
approximation to $\partial^2 l(\theta)/\partial \theta_k^2$ which could have a
detrimental effect on the convergence of the coordinate descent algorithm. An
important observation is
that~\eqref{eq:coord_update}--\eqref{eq:coord_update:st} used with an
approximated second-order derivative still yield the correct stationary points.

To see why it is true, assume that $\bar{\theta}$ is such
that~\eqref{eq:coord_update}--\eqref{eq:coord_update:st} leave $\bar{\theta}_k$
unchanged (i.e., $\theta_k = \bar{\theta}_k$). If $\bar{\theta}_k = 0$, this can
happen only if $|\partial l(\bar{\theta})/\partial \theta_k| \leq \rho_1$, which
is indeed the first order optimality condition in 0. Now assume that
$\bar{\theta}_k > 0$, the fact that $\bar{\theta}_k$ is left unmodified by the
recursion implies that $\bar{\theta}_k \rho_2 + \partial l(\theta)/\partial
\theta_k + \rho_1 = 0$, which is also recognized as the first order optimality
condition (note that since $\bar{\theta}_k \neq 0$, the criterion is
differentiable at this point). The symmetric case, where $\bar{\theta}_k
< 0$, is similar. Hence, the use of an approximated second-order
derivative does not prevent the algorithm from converging to the appropriate
solution. A more subtle issue is the question of stability: it is easily checked
that if $\partial^2 l(\theta)/\partial \theta_k^2$ is smaller than it should be
(remember that it has to be positive as $l(\theta)$ is strictly convex), the
algorithm can fail to converge even for simple functions (e.g., if $l(\theta)$
is a quadratic function). An elaborate solution to this issue would consist in
performing a line search in the ``direction'':
\[
  \frac{s\left(\alpha^{-1} \frac{\partial^2 l( \bar{\theta})}{\partial \theta_k^2} \bar{\theta}_k - \frac{\partial l( \bar{\theta})}{\partial \theta_k}, \rho_1\right)}{\alpha^{-1} \frac{\partial^2 l( \bar{\theta})}{\partial \theta_k^2} + \rho_2}
\]
where $0 < \alpha \leq 1$, is chosen as close as possible to 1 with the
constraint that it indeed leads to a decrease of the objective function (note
that the step size affects only the second-order term in order to preserve the
convergence behavior). On the other hand, coordinate descent algorithms are
only viable if each individual update can be performed very quickly, which means that
using line search is not really an option. In our experiments, we found that
using a fixed value of $\alpha=1$ was sufficient for Algorithm~\ref{alg:crfcoord},
probably due to the fact that the second-order derivative approximation
in~(\ref{eq:2ndorder_approx}) is usually quite good. For the blockwise approach
described below, we had to use somewhat larger values of $\alpha$ to ensure stability
(typically, in the range 2--5 near convergence and in the range 50--500 for the very
few initial steps of the algorithm in cases where it is started blindly from arbitrary
parameter values). Alternative updates based on uniform upper-bounds of the Hessian could
also be derived in a fashion similar to the work reported in \cite{Krishnapuram05sparse}.

\subsection{Blockwise Coordinate Descent for CRFs}
The algorithm described in the previous section is efficient in simple problems
(see Section~\ref{sec:simul}) but cannot be used even for moderate size
applications of CRFs. For instance, the application to be considered in
Section~\ref{sec:exper} involves up to millions of parameters and single component
coordinate descent is definitely ruled out in this case. Following 
\cite{Friedman08GLM}, we investigate the use of blockwise updating schemes,
which update several parameters simultaneously trying to share as much
computations as possible. It turns out that the case of CRFs is rather
different from the polytomous logistic regression case considered
in~\cite{Friedman08GLM} and requires specific blocking schemes. In this discussion,
we consider the parameterization defined in (\ref{eq:mu_lambda}) which 
makes it easier to highlight the proposed block structure.

Examining the forward-backward procedure described in Section~\ref{sec:fb}
shows that the evaluation of the first or second order derivative of the
objective function with respect to $\mu_{y,x}$ or $\lambda_{y',y,x}$ requires
to compute the pairwise probabilities $\operatorname{p}_\theta(y_{t} = y'',
y_{t+1} = y' | \mathbf{x}^{(i)})$ \emph{for all values of $(y'',y') \in Y^2$}
and for all sequences $\mathbf{x}^{(i)}$ which contain the symbol $x$ at any
position in the sequence. Hence, the most natural grouping in this context is
to simultaneously update all parameters
$\{\mu_{y,x},\lambda_{y',y,x}\}_{(y',y)\in Y^2}$ that correspond to the same
value of $x$. This grouping is orthogonal to the solution adopted for
polytomous regression in~\cite{Friedman08GLM}, where parameters are
grouped by common values of the target label.


\begin{algorithm}
\caption{Blockwise Coordinate Descent for CRF}
\label{alg:block_crfcoord}
\begin{algorithmic}
\WHILE {Convergence criterion is not met}
\FOR {$x \in X$}
\FOR {Sequences which contain the symbol $x$}
\STATE Perform sparse forward-backward on relevant indices.
\ENDFOR
\STATE Compute 
\begin{align*}
  & \{\partial l (\mu,\lambda)/\partial \mu_{y,x}, \partial^2
l(\mu,\lambda)/\partial \mu_{y,x}^2\}_{y\in Y} \\
  & \{\partial l (\mu,\lambda)/\partial
\lambda_{y',y,x}, \partial^2 l(\mu,\lambda)/\partial \lambda_{y',y,x}^2\}_{(y',y)\in Y^2}
\end{align*}
using~(\ref{eq:1rstorder}) and~\eqref{eq:2ndorder_approx}.
\STATE Update $\{\mu_{y,x}\}_{y\in Y}$ and then $\{\lambda_{y',y,x}\}_{(y',y)\in Y^2}$ according to~\eqref{eq:coord_update}--\eqref{eq:coord_update:st}.
\ENDFOR
\ENDWHILE
\end{algorithmic}
\end{algorithm}

Different variants of this algorithm are possible, including updating only one of the two types of
blocks ($\{\mu_{y,x}\}_{y\in Y}$ or $\{\lambda_{y',y,x}\}_{(y',y)\in Y^2}$) at a time.
Although the block coordinate-wise algorithm requires scanning all the $|X|$
possible symbols $x$ at each iteration, it is usually relatively fast due to
the fact that only those sequences that contain $x$ are considered. In
addition, careful examination reveals that for each sequence that contains the
token $x$, it is only required to carry the forward recursion up to the index of
last occurrence of $x$ in the sequence (and likewise to perform the backward
recursion down to the first occurrence of $x$).
The exact computational saving will 
however depend on the target application as discussed in Section~\ref{ssec:nettalk} below.


\section{Discussion}
\label{sec:discus}

As mentioned above, the standard approach for CRFs is based on the use of the $\ell^2$ penalty term
and the objective function is optimized using L-BFGS \cite{Nocedal89lbfgs}, conjugate gradient
\cite{nodecal-wright06book} or Stochastic Gradient Descent (SGD) \cite{bottou04mlss}. The CRF
training softwares CRF++ \cite{kudo05crfpp} and CRFsuite \cite{okazaki07crfsuite} use L-BFGS while
crfsgd \cite{bottou07sgd} is, as the name suggests, based on SGD. The latter approach differs from
the others in that it processes the training sequences one by one: thus each iteration of the
algorithm is very fast and it is generally observed that SGD converges faster to the solution,
especially for large training corpora.
On the other hand, as the algorithm approaches convergence, SGD becomes slower than
global quasi-Newton algorithms such as L-BFGS. \cite{Vishwanathan06accelerated} discusses
improvements of the SGD algorithm based on the use of an adaptive step size whose computation
necessitates second-order information. However, these approaches based on the $\ell^2$ penalty term
do not perform feature selection.

To our knowledge, \cite{McCallum03features} made the first attempt to perform model selection for
conditional random fields. The approach was mainly motivated by \cite{DellaPietra97inducing} and is
based on a greedy algorithm which selects features with respect to their impact on the
log-likelihood function. Related ideas also appear in \cite{Dietterich04gradientboost}. These
greedy approaches are different from our proposed algorithm in that they do not rely on a
convex optimization formulation of the learning objective.

To deal with $\ell^1$ penalties, the simplest idea is that of \cite{Kazama03evaluation} which was
introduced for maximum entropy models but can be directly applied to conditional random fields. The
main idea of \cite{Kazama03evaluation} is to split every parameter $\theta$ into two positive
constrained parameters, $\theta^+$ and $\theta^-$, such that $\theta = \theta^+ - \theta^-$. The
penalty takes the form $\rho(\theta^+ - \theta^-)$. The optimization procedure is quite simple, but
the number of parameters is doubled and the method is reported to have a slow convergence
\cite{Andrew07L1Reg}. A more efficient approach is the already mentioned orthant-wise quasi-Newton algorithm
introduced in \cite{Andrew07L1Reg}.
\cite{Gao07comparativeNLP} shows that the orthant-wise
optimization procedure is faster than the algorithm proposed by \cite{Kazama03evaluation} and performs
model selection even in very high-dimensional problems with no loss of performance compared to
the use of the $\ell^2$ penalty term. Orthant-wise optimization is available in the CRFsuite
\cite{okazaki07crfsuite} package. Recently, \cite{Tsuruoka09stochastic} proposed an
adapted version of SGD with $\ell^1$ penalization, which is claimed to be much faster than
the orthant-wise algorithm.

As observed in \cite{Peng06information,Gao07comparativeNLP}, $\ell^1$ regularization per se does
not in general warrant improved test set performance. We believe that the real challenge is to come
up with methods for CRFs that can take profit of the parameter sparsity to either speed up
processing or, more importantly, make it possible to handle larger ``virtual'' sets of parameters
(i.e., a number of parameters that is potentially very large but only a very limited fraction of
them being selected). The combined contributions of Sections~\ref{sec:inference}
and~\ref{sec:estimation} are a first step in that direction. Related ideas may be found in
\cite{Cohn06efficient} who considers ``generalized'' feature functions. Rather than making each
feature function depend on a specific value of the label (or on specific values of label pairs),
the author introduces functions that only depend on subsets of (pairs of) labels. This amounts to
introducing tying between some parameter values, a property that can also be used to speed-up the
forward-backward procedure during training. This technique allows to considerably reduce the
training time, with virtually no loss in accuracy. From an algorithmic perspective, this work is
closest to our approach, since it relies on a decomposition of the clique potential into two terms,
the first has a linear complexity (w.r.t.\ the number of labels), and the other is sparse; this idea
was already present in \cite{Siddiqi05fast}. This method however requires to specify a priori the
tying pattern, a requirement that is not needed here. The important dependencies emerge from the
data, rather than being heuristically selected a priori. A somewhat extreme position is finally
advocated in \cite{Liang08trading}, where the authors propose to trade the explicit modeling of
dependencies between labels for an increase in the number of features testing the local
neighborhood of the current observation token. Our proposal explores the opposite choice: reducing
the number of features to allow for a better modeling of dependencies.

\section{Experiments}
\label{sec:exper}

\subsection{Simulated data}
\label{sec:simul}
The experiments on an artificial dataset reported here are meant to illustrate two aspects of
the proposed approach. Firstly, we wish to show that considering unnecessary dependencies in a model can
really hurt the performance, and that using $\ell^2$ and $\ell^1$
regularization terms can help solve this problem. Second, we wish to demonstrate
that the blockwise algorithm (Algorithm~\ref{alg:block_crfcoord}) enables to achieve accuracy
results that are very close to those obtained with the coordinate descent approach
(Algorithm~\ref{alg:crfcoord}).

The data we use for these experiments are generated with a first-order hidden Markov model. Each
observation and label sequence has a length of $5$, the observation alphabet contains $5$ values,
and the label alphabet contains $6$ symbols. This HMM is designed in such a way that the transition
probabilities are uniform for all label pairs, except for two. The emission probability matrix on
the other hand has six distinctively dominant entries such that most labels are well identified from
the observations, except for two of them which are very ambiguous. The minimal (Bayes) error for
this model is $15.4 \%$.

\begin{figure}
\centering
\includegraphics[width=0.35\textwidth]{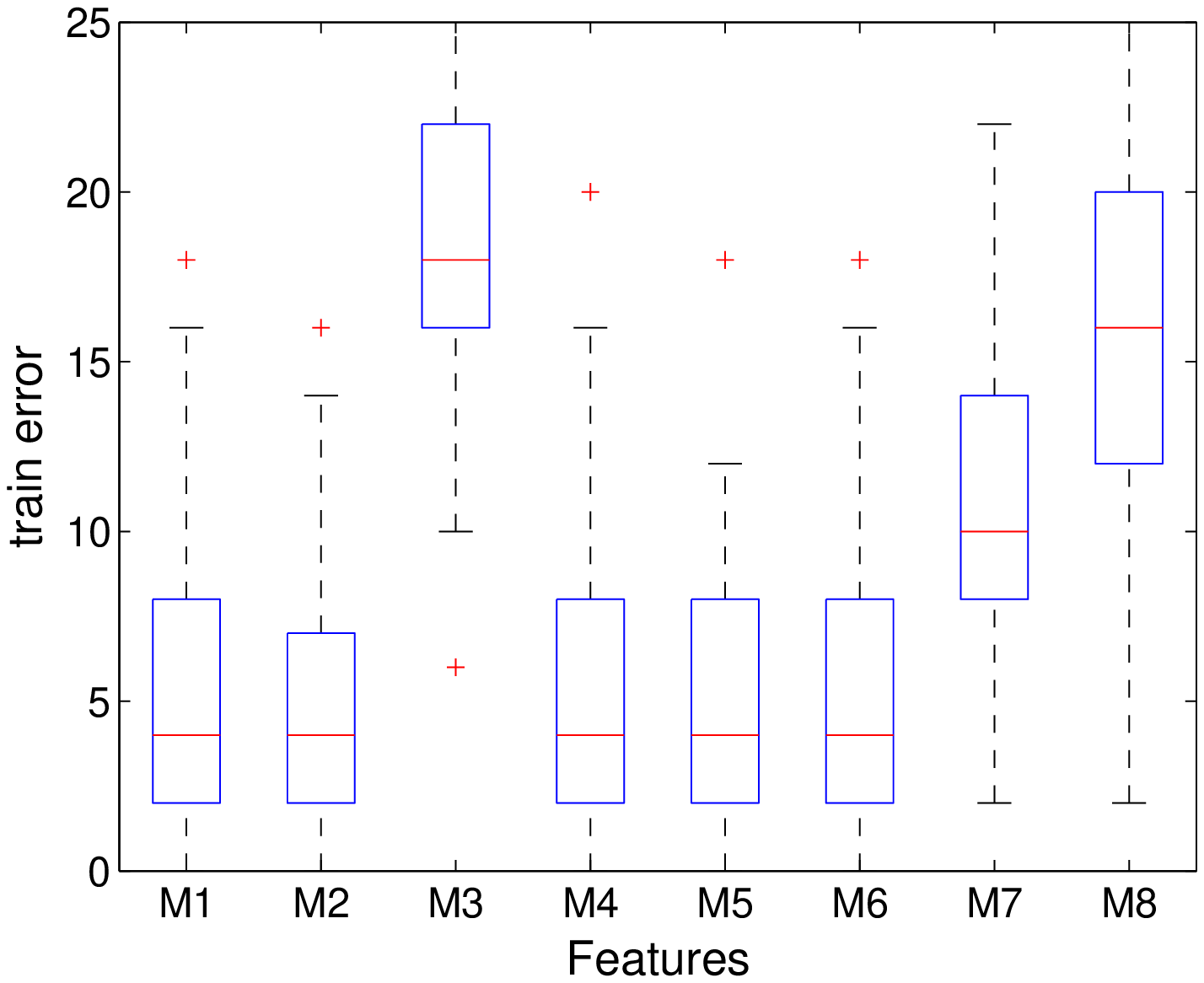} \qquad \includegraphics[width=0.35\textwidth]{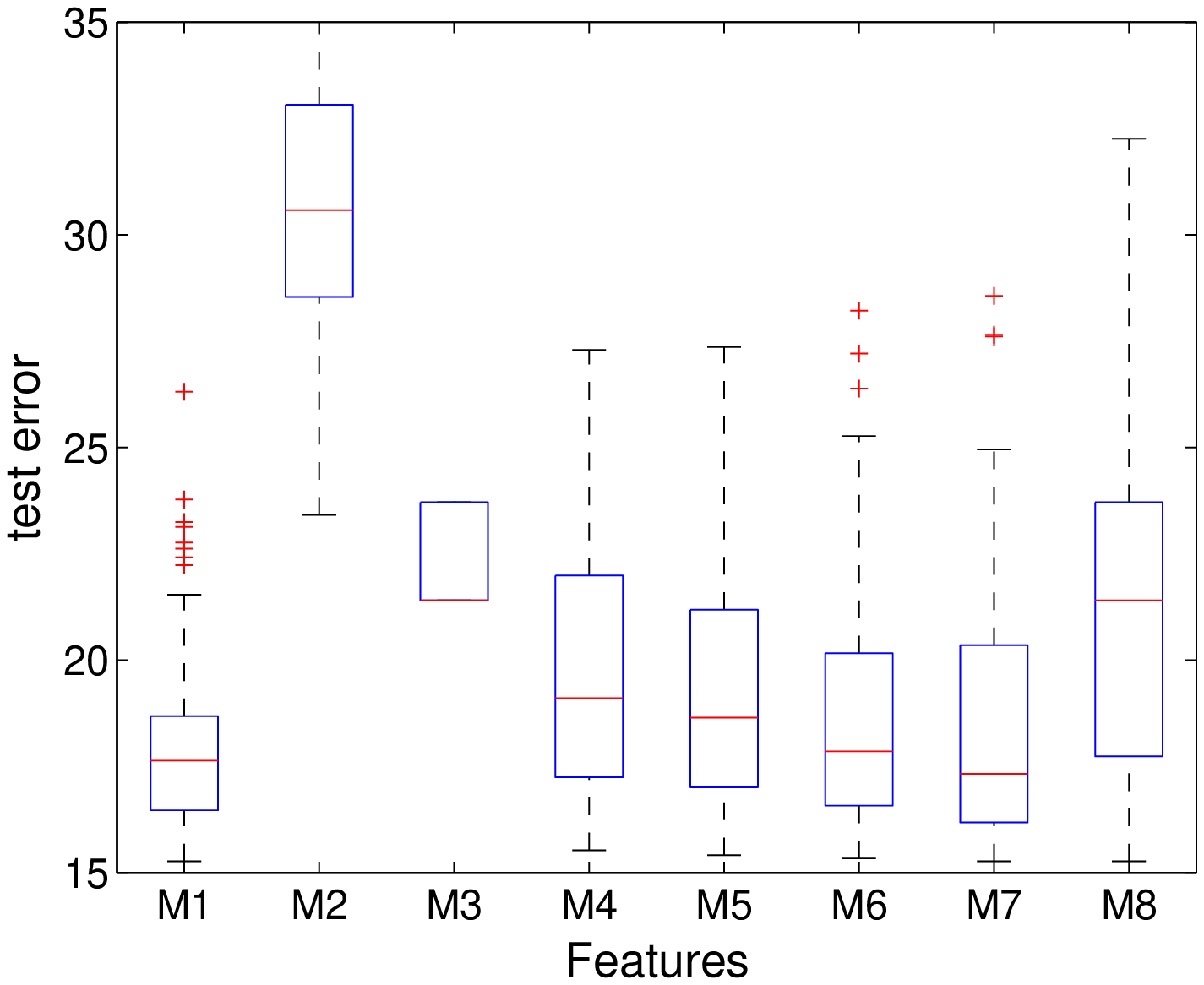}
\caption{Performance of the models on artificial data. Models $M1-M3$ are trained with $\ell^2$ penalty (L-BFGS), models $M4-M8$ with the $\ell^1$ penalty term (block coordinate-wise descent). Above: performance on training set. Below: performance on testing set.}
\label{fig:simulatedlongdep}
\end{figure}

Figure~\ref{fig:simulatedlongdep} compares several models: M1 contains both $(y_{t-1}, y_t, x_t)$
and $(y_t, x_t)$ features, M2 and M3 are simpler, with M2 containing only the bigram features, and
M3 only the unigram features. The models M1--M3 are penalized with the $\ell^2$ norm. Models M4--M8
contain both features, bigram and unigram, but are penalized by the elastic net penalty term. For
the $\ell^2$-penalized models (M1--M3), the regularization factor $\rho_2$ is set to its optimal
value (obtained by cross validation). For M4--M8 however, the value of $\rho_2$ does not influence
much the performance and is set to $0.001$ while M4--M8 correspond to different choices of $\rho_1$,
as shown in Table \ref{simulated_lasso}.

For this experiment, we used only $N=10$ sequences for training, so as to reproduce the situation,
which is prevalent in practical uses of CRFs, where the number of training tokens (here $10 \times
5 = 50$) is of the same order as the number of parameters, which ranges from $6 \times 5 = 30$ for
M3 to $6 \times 5 + 6^2 \times 5 = 210$ for M1 and M4--M8. Figure~\ref{fig:simulatedlongdep}
displays box-and-whiskers plots summarizing 100 independent replications of the experiment.

\begin{table}[h!]
\centering
\begin{tabular}{|c|c|c|c|c|c|}
\hline
& M4 & M5 & M6 & M7 & M8 \\
\hline
$\rho_1$ & 0.001 & 0.01 & 0.1 & 1 & 2.5 \\
Number of active $\mu$ & 28.5 & 15.0 & 10.9 & 6.2 & 5.8 \\
Number of active $\lambda$ & 50.6 & 26 & 17.2 & 4.9 & 1.3 \\
\hline
\end{tabular}
\caption{Impact of $\rho_1$ on the number of active features ($\rho_2 = 0.001$).}
\label{simulated_lasso}
\end{table}

Unsurprisingly, M1 and M2, which contain more parameters, perform very well on the training set,
much better than M3. The test performance tells a different story: M2 performs in fact much worse
that the simple unigram model M3, which is all the more remarkable that we know from the simulation
model that the observed tokens are indeed not independent and that the models are nested (i.e. any
model of type M3 corresponds to a model of type M2). Thus, even with regularization, richer models
are not necessary the best, hence the need for feature selection techniques. Interestingly, M1
which embarks both unigram and bigram features, achieves the lowest test error, highlighting the
interest of using simultaneously both feature types to achieve some sort of smoothing effect. With
proper choice of the regularization (here, M7), $\ell^1$-penalized models achieve comparable
test set performance. As a side effect of model selection, notice that M7 is somewhat better than
M1 at predicting the test performance at training time: for M1, the average train error is 6.4\%
vs. 18.5\% for the test error while for M7, the corresponding figures are 10.3\% and 17.9\%,
respectively. Finally, closer inspection of the sparsity pattern determined by M7 shows that it is
most often closely related to the structure of the simulation model which is also encouraging.
 
\begin{figure}[hbt]
\centering
\includegraphics[width=0.35\textwidth]{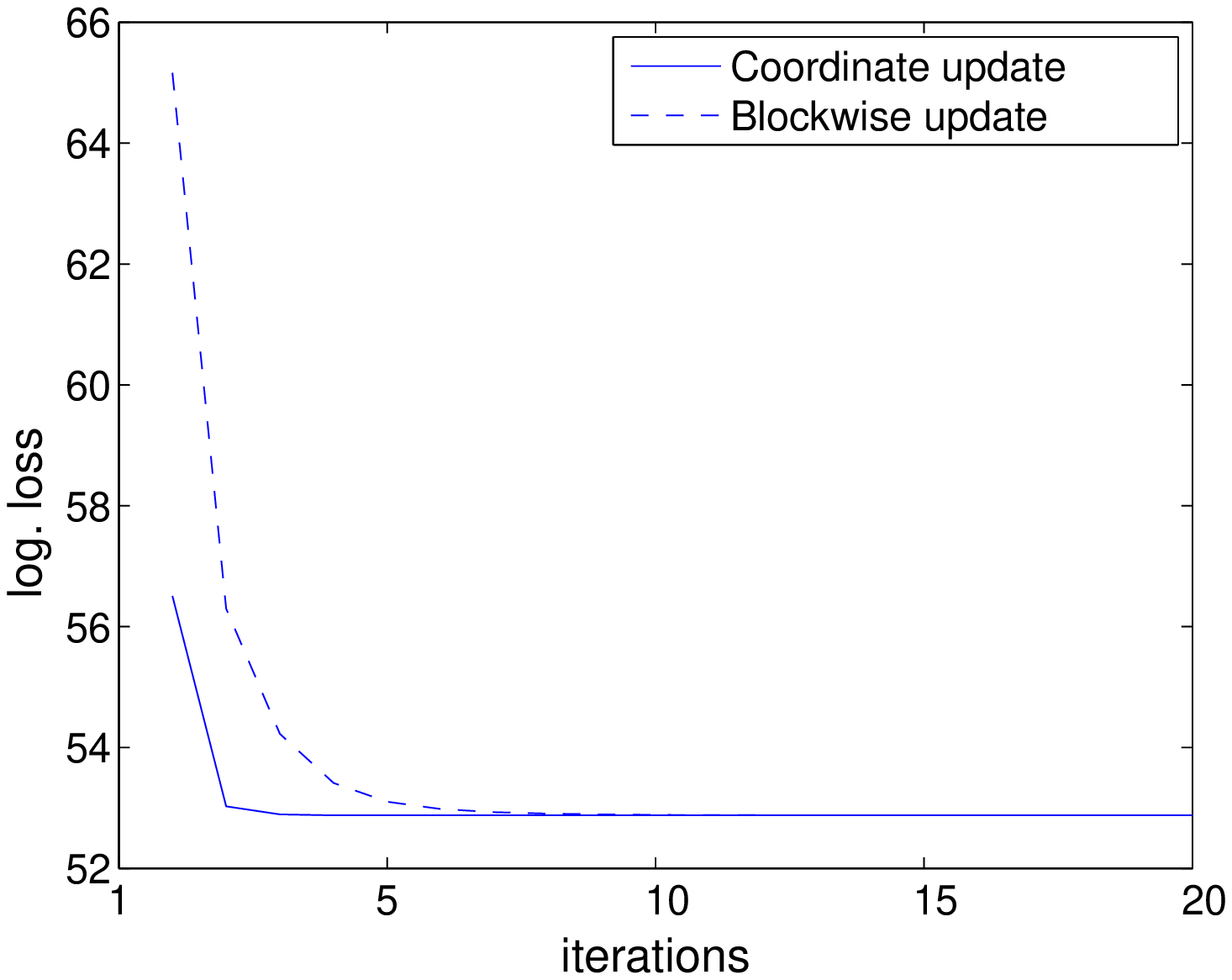} \qquad \includegraphics[width=0.35\textwidth]{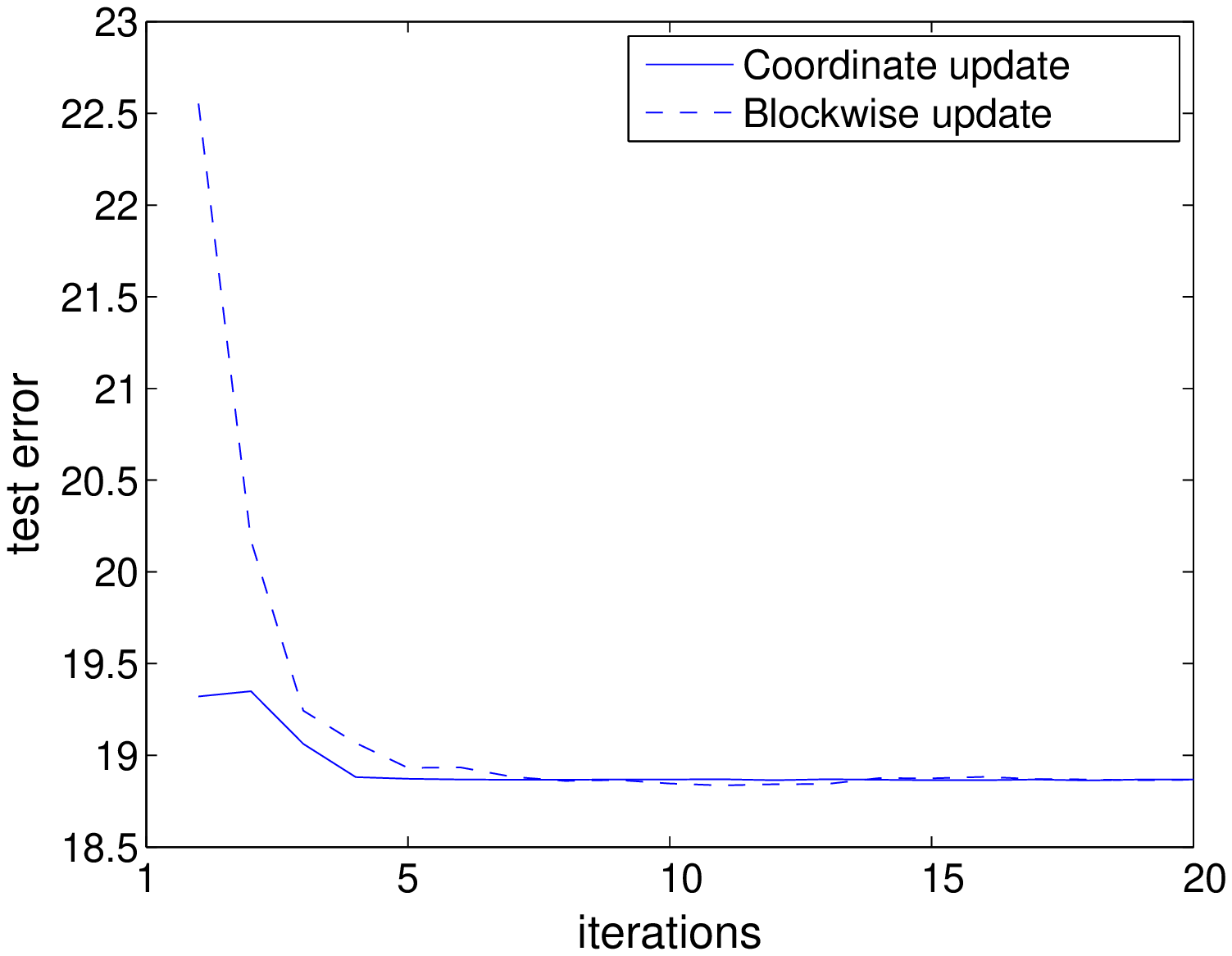} 
\caption{Comparison of the coordinate-wise update with the block update on simulated data. Above: values of logarithmic loss being minimized. Below: performance on test data.}
\label{fig:twoalgorithms}
\end{figure}

Figure~\ref{fig:twoalgorithms} compares the behavior of the coordinate-wise update policy with the
blockwise approach, where one iteration refers to a complete round where all model parameters are
updated exactly once. As can be seen on these graphs, the convergence behavior is comparable for
both approaches, both in terms of objective function (top plot) and test error (bottom plot). Each
iteration of the blockwise algorithm is however about 50 times faster than the coordinate-wise
update, that roughly correspond to the size of each block. Clearly, the blockwise approach is the
only viable strategy when tackling more realistic higher-dimensional tasks such as those considered
in the next two sections.

\subsection{Experiments with Nettalk}
\label{ssec:nettalk}
This section presents results obtained on a word phonetization task, using the
Nettalk dictionary~\cite{Sejnowski87}. This dataset contains approximately
18000 English words and their pronunciations. Graphemes and phonemes are
aligned at the character level thanks to the insertion of a ``null sound'' in
the latter string when it is shorter than the former.  The set of graphemes $X$
thus includes $26$ letters, the alphabet of phonemes $Y$ contains $53$
symbols. In our experiments, we consider that each phoneme is a target
label and we consider two different settings. The first only uses features that test the value of one single
letter, and is intended to allow for a detailed analysis of the
features that are extracted. The second setting is more oriented
towards performance and uses features that also test the neighboring letters.
The training set comprises $16452$ sequences and the test set contains $1628$ sequences. The
results reported here are obtained using the blockwise version of the coordinate
descent approach (Algorithm~\ref{alg:block_crfcoord}).

\begin{figure}[hbt]
\centering
\includegraphics[width=0.35\textwidth]{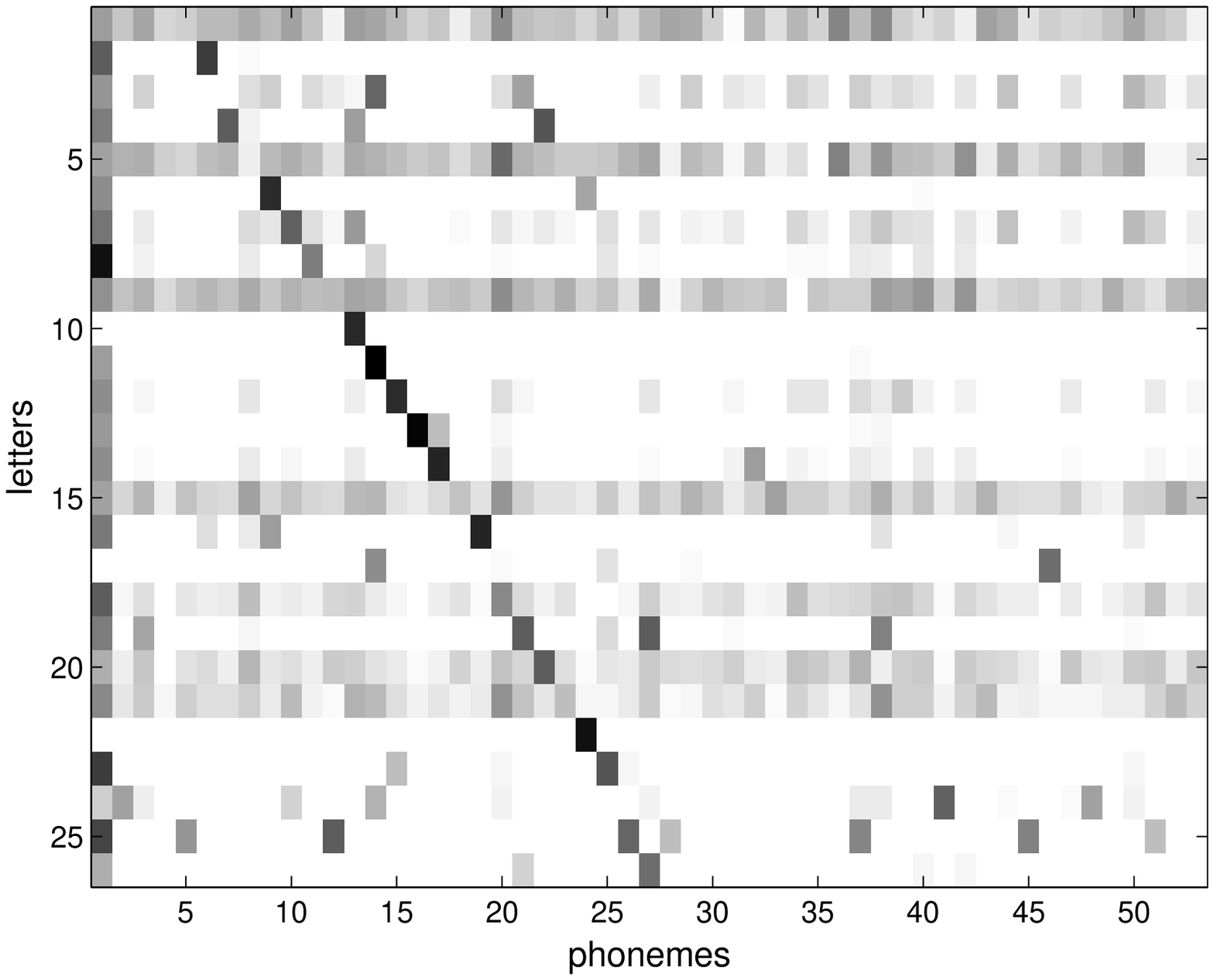} \qquad \includegraphics[width=0.35\textwidth]{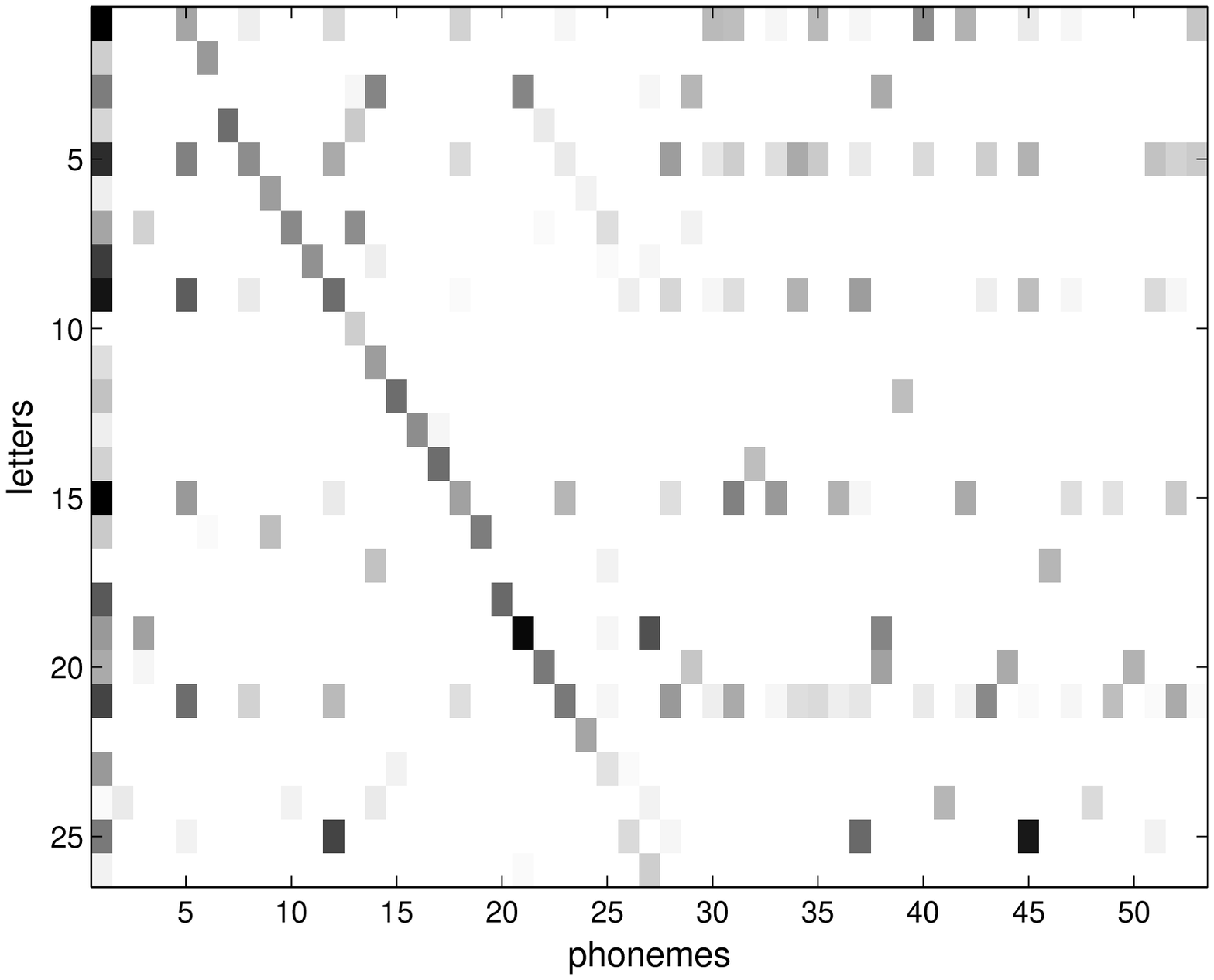}
\caption{Nettalk task, $\ell^1$ norm of the parameters estimated with elastic net penalty, $\rho_1=0.2$, $\rho_2=0.05$ Above: $|\mu_{y,x}|$ for the  53 phonemes $y$ and 26 letters $x$. Below: $\sum_{y'} |\lambda_{y',y,x}|$ for the 53 phonemes $y$ and 26 letters $x$.}
\label{fig:crfl1_nettalk}
\end{figure}

Figure~\ref{fig:crfl1_nettalk} displays the estimated parameter values when the $\ell^1$ penalty is
set to its optimal value of $0.2$ (see Table~\ref{tab:crfl1_nettalk} below). Comparing this figure
with Figure~\ref{fig:crfl2_nettalk} shows that the proposed algorithm correctly identifies those
parameters that are important for the task while setting the other to zero. It confirms the
impression conveyed by Figure~\ref{fig:crfl2_nettalk} that only a very limited number of features is
actually used for predicting the labels. The first column in this
figure corresponds to the null sound and is unsurprisingly associated with almost
all letters. One can also directly visualize the ambiguity of the vocalic graphemes which
correspond to the first ('a'), fifth ('e'), ninth ('i'), etc. gray rows; this contrasts with the
much more deterministic association of a typical consonant grapheme with a single consonant
phoneme.

\begin{table}[hbt]
  \centering
  \begin{tabular}{|c|c|c|c|c|c|c|} \hline
    Method & Iter. & Time & Train & Test & $K_\mu$ & $K_\lambda$\\
      &    &  (min.) & (\%) &   (\%)  & & \\ \hline
    SBCD,  $\rho_1 = 0$   & 30 & 125 & 13.3 & 14.0  & 1378 & 73034 \\ 
    SBCD,  $\rho_1 = 0.1$ & 30 & 76 & 13.5 & 14.2 & 1155 & 4171   \\
    SBCD,  $\rho_1 = 0.2$ & 30 & 70 & 14 & 14.2 &  1089 & 3598   \\
    SBCD,  $\rho_1 = 0.5$ & 30 & 63  & 13.7 & 14.3 & 957 & 3077   \\
    SBCD,  $\rho_1 = 1$   & 30 & 55 & 16.3 & 16.8 &  858 & 3111     \\
    SBCD,  $\rho_1 = 2$   & 30 & 43 & 16.4 & 16.9 &  760 & 2275     \\
    SBCD,  $\rho_1 = 10$  & 30 & 25 & 17.3 & 17.7 & 267 & 997	   \\ 
    \hline 
	 OWL-QN, $\rho_1 = 0.1$ & 50 & 165 & 13.5 & 14.2  & 1864 & 4079 \\
    \hline\hline
    L-BFGS & 90 & 302 & 13.5 & 14.1  & \multicolumn{2}{c|}{74412}  \\
    \hline
    SGD & 30 & 17 & 18.5 & 19.1  & \multicolumn{2}{c|}{74412}  \\
    \hline
  \end{tabular}
  \caption{Upper part: summary of results for various values of $\rho_1$ for the proposed Sparse Blockwise Coordinate Descent (SBCD) algorithm (with $\rho_2 = 0.001$) and orthant-wise L-BFGS (OWL-QN). Lower part: results obtained with $\ell^2$ regularization only, for L-BFGS and stochastic gradient descent (SGD).}
  \label{tab:crfl1_nettalk}
\end{table}

Table~\ref{tab:crfl1_nettalk} gives the per phoneme accuracy with varying level of sparsity, both
for the proposed algorithm (SBCD) and the orthant-wise L-BFGS (OWL-QN) strategy of
\cite{Andrew07L1Reg}. For comparison purposes the lower part of the table also reports performance
obtained with $\ell^2$ regularization only. For $\ell^2$-based methods (L-BFGS and SGD) the
regularization constant was set to its optimal value determined by cross validation as $\rho_2 =
0.02$. The proposed algorithm (SBCD) is coded in C while OWL-QN and L-BFGS use the CRF++ package
\cite{kudo05crfpp} modified to use the \verb+liblbfgs+ library provided with CRFsuite
\cite{okazaki07crfsuite} that implements the standard and orthant-wise modified versions of
L-BFGS. Finally, SGD use the software of \cite{bottou07sgd}. All running times were measured on a
computer with an Intel Pentium 4 3.00GHz CPU and 2G of RAM
memory. Measuring running times is a difficult issue as each iteration
of the various algorithm does not achieve the same improvement in
term of performance. For our method, 30 iterations were found necessary to reach
reasonable performance in the sense that further iterations did not significantly reduce the error
rates (with variations smaller than 0.3\%). Proceeding similarly for the other methods showed that
OWL-QN and L-BFGS usually require more iterations to reach stable performance, which is reflected
in Table~\ref{tab:crfl1_nettalk}. Finally, SGD requires few iterations (where an iteration is
defined as a complete scan of all the training sequences) although we obtained disappointing
performance on this dataset with SGD.

First, Table~\ref{tab:crfl1_nettalk} shows that for $\rho_1 = 0.1$ or $0.2$ our method
reaches an accuracy that is comparable with that of non-sparse trainers (SBCD with $\rho_1=0$ or
L-BFGS) but with only about 5000 active features. Note in particular the dramatic reduction
achieved for the bigram features $\lambda_{y',y',x}$ as the best accuracy/sparsity compromise
($\rho_2 = 0.2$) nullifies about 95\% of these parameters. We observe that the performance of SBCD
(for $\rho_1 = 0.1$) is comparable to that of OWL-QN, which is reassuring as they optimize related
criterions, except for the fact that OWL-QN is based on the use of the sole $\ell^1$ penalty. There
are however minor differences in the number of selected features for both methods. In addition to
the slight difference in the penalties used by SBCD and OWL-QN, it was
constantly observed in all our experiments that for $\ell^1$-regularized methods the performance
stabilizes much faster than the pattern of selected features which may require as much as a few
hundreds of iterations to fully stabilize. This effect was particularly noticeable with the OWL-QN
algorithm. We have not found satisfactory explanation regarding the poor performance of SGD on this
dataset: further iterations do not significantly improve the situation and this failure has not
been observed on the CoNLL 2003 data considered below. In general, SGD is initially very fast to
converge and no other algorithm is able to obtain similar performance with such small running
time. The fact that SGD fails to reach satisfactory performance in this example is probably related
to an incorrect decrease of the step size. In this regard, an important difference between the
Nettalk data and the CoNLL 2003 example considered below is the number of possible labels which is
quite high here (53). A final remark regarding timings is that all methods except SBCD use
logarithmic computation in the forward-backward recursions. As discussed in Section~\ref{sec:fb},
this option is slower by a factor which, in our implementation, was measured to be about
$2.4$. Still, the SBCD algorithm compares favorably with other algorithms, especially with OWL-QN
which optimizes the same objective function.

\begin{figure}[hbt]
\centering
\includegraphics[width=0.45\textwidth]{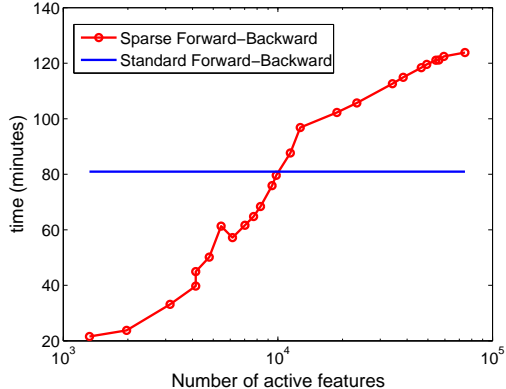}  
\caption{Running time as a function of the number of active features for the SBCD algorithm on the Nettalk corpus. The blue line correspond to the running time when using non-sparse forward-backward.}
\label{fig:nettalktime}
\end{figure}

Table~\ref{tab:crfl1_nettalk} also shows that the running time of SBCD depends on the
sparsity of the estimated model, which is fully attributable to the sparse version of the
forward-backward recursion introduced in Section~\ref{sec:fb}. To make this connection
clearer, Figure~\ref{fig:nettalktime} displays the running time as a function of the
number of active features (rather than $\rho_1$). When the number of active feature is
less than 10000, the curve shows a decrease that is proportional to the number of active
features (beware that the x-axis is drawn on a logarithmic scale). The behavior observed
for larger numbers of active features, where the sparse implementation becomes worse than
the baseline (horizontal blue line) can be attributed to the overhead generated by the use
of sparse matrix-vector multiplications for matrices that are not sparse. Hence the ideas
exposed in Section~\ref{sec:fb} have a strong potential for reducing the computational
burden in situations where the active parameter set is very small compared to the total
number of available features. Note also that the OWL-QN optimizer could benefit from this
idea as well.

The simple feature set used in the above experiments is too restricted to achieve
state-of-the-art performance for this task. We therefore conducted another series of
experiments with much larger feature sets, including bias terms and tests on the neighbors of the letter
under focus.  For these experiments, we keep the same baseline feature set and add the
bias terms $\1(y_t=y)$ and $\1(y_{t-1}=y', y_t=y)$ for all possible values of
$(y,y')$. For the context, we consider two variants: in the first, termed
\emph{pseudo n-gram}, we also add $\1(y_t=y, x_{t\pm{}i}=x)$ and $\1(y_{t-1}=y',y_t=y,
x_{t\pm{}i}=x)$ for all values of $(x, y,y')$ and of the offset $1 \leq i \leq
n-1/2$. In other terms, in this variant, we test separately the values of the
letters that occur before and after the current position.  In the second variant, termed
\emph{n-gram}, we add features $1(y_t=y, z_t=z)$ and $1(y_{t-1}=y',y_t=y, z_t=z)$ where $z_t$
denotes the letter k-gram centered on $x_t$, and $z$ ranges over all observed
k-grams (with $k \leq n$). This second variant seems of course much more computationally demanding
\cite{Liang08trading} as it yields a much higher number of features (see the top line of
Table~\ref{tab.nettalk2}, where the total number of features is given in millions).

\begin{table}[hbtp]
  \centering
  \begin{tabular}{|l|c|c||c|c|} \hline
    & \multicolumn{2}{c||}{pseudo n-gram}    & \multicolumn{2}{c|}{n-gram} \\
    & $n=3$ & $n=5$    & $n=3$ & $n=5$ \\ 
   M feat. & 0.236 & 0.399 & 14.2 & 121 \\ \hline \hline
   20 iter.   & 8.98 (12.5) & 6.77 (19.8) & 8.22 (12.8) & 6.51 (21.7) \\
   30 iter.   & 8.67 (10.9) & 6.65 (17.1) & 8.04 (11.7) & 6.50 (20.1) \\ \hline
  \end{tabular}
  \caption{Experiments with contextual features: performance of the SBCD algorithm after
    20 and 30 iterations in terms of error rate and, between parentheses, number of
    selected features (in thousands).} 
 \label{tab.nettalk2}
\end{table}

As can be seen in Table~\ref{tab.nettalk2}, these extended feature sets yield
results that compare favorably with those reported in \cite{Yvon96overlapping,Marchand00multi}, 
with an phoneme-error-rate of 6.5\% for the 5-gram system. We also find that even though
both variants extract comparable numbers of features, the results
achieved with ``true'' n-grams are systematically better than for the pseudo n-grams. More
interestingly, the n-gram variants are also faster: this paradoxical observation is due to
the fact that for the n-gram features, each block update only visits a very small number
of observation sequences, and further, that for each position, a much smaller number of
features are active as compared to the pseudo n-gram case.  Finally, the analysis of the
performance achieved after 20 and 30 iterations suggests that the n-gram systems are
quicker to reach their optimal performance. This is because a very large proportion of
the n-gram features are zeroed in the first few iterations. For instance, the 5-gram
model, after only 9 iterations, has an error rate of 6.56\% and selects only 27.3 thousand
features out of 121 millions. These results clearly demonstrate the computational reward
of exploiting the sparsity of the parameter set as described in
Section~\ref{sec:inference}: in fact, training our largest model takes less than 5 hours (for
20 iterations), which is quite remarkable given the very high number of features.

\subsection{Experiments with CoNLL 2003}
\label{ssec:conll03}

Named entity recognition consists in extracting groups of syntagms that correspond to named
entities (e.g., names of persons, organizations, places, etc.). The data used for our experiments
are taken from the CoNLL 2003 challenge \cite{Tjong03introduction} and implies four distinct types
of named entities, and $8$ labels. Labels have the form B-X or I-X, that is begin or inside of a
named entity X (however, the label B-PER is not present in the corpus). Words that are not included
in any named entity, are labeled with O (outside). The train set contains 14987 sequences, and the
test set $3684$ sequences. At each position in the text, the input consists of three separate
components corresponding respectively to the word (with $30290$ distinct words in the corpus),
part-of-speech ($46$), and syntactic ($18$) tags. To accommodate this multidimensional input the
standard practice consists in superposing unigram or bigram features corresponding to each of those
three dimensions considered separately. The parameters we use in the model are of the form
$\{\mu_{y,x^d} \, , \lambda_{y',y, x^d}\}$ for $d \in \{1,2,3\}$, which corresponds to a little more than
9 million parameters. Hence, the necessity to perform model selection is acute.
 
\begin{figure}[hbt]
\centering
 \includegraphics[width=0.45\textwidth]{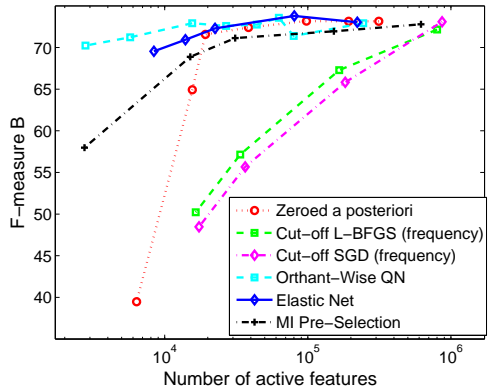}
\caption{Performance comparison of different model selection approaches on CoNLL 2003 (English): test set B.}
\label{fig:conll03}
\end{figure}

To illustrate the efficiency of $\ell^1$-based feature selection, we compare it with three
simple-minded approaches to feature selection, which are often used in practice. The first
one, termed ``cut-off'', consists in incorporating only those features that have been
observed sufficiently often in the training corpus. This amounts to deleting \emph{a
  priori} all the rare dependencies. The second strategy, termed \emph{MI preselection},
selects features based on their Mutual Information (MI) with the label, as in
\cite{Yang97comparative} (where the MI is referred to as the \emph{information gain}). 
The third option consists in training a model that is not sparse
(e.g., with an $\ell^2$ penalty term) and in removing \emph{a posteriori} all features whose
values are not of sufficient magnitude. Figure \ref{fig:conll03} compares the error rates
obtained with these strategies to those achieved by the SBCD or OWL-QN
algorithms. Obviously, a priori cut-off strategies imply some performance degradation, although MI
preselection clearly dominates frequency-based selection.
A posteriori thresholding is more
efficient but cannot be used to obtain well-performing models that are very sparse (here,
with less than 10000 features). From a computational point of view, a posteriori thresholding
is also penalized by the need to estimate a very large model that contains all available features.


In this experiment, SBCD proves computationally less efficient, for at
least two reasons. First, many POS or chunk tags appear in all training
sequences; the same is true for very frequent words: this yields many
very large blocks containing almost all training sentences. Second, the sparse
forward-backward implementation is less efficient than in the case of the phonetization task as the
number of labels is much smaller: SBCD needs 42 minutes (with $\rho_1 = 1$, corresponding to 6656
actives features) to achieve a reasonable performance while OWL-QN is faster, taking about 5
minutes to converge. If sparsity is not needed, SGD appears to be the most efficient method for
this corpus as it converges in less than 4 minutes. L-BFGS in contrast requires about 25 minutes to
reach a similar performance.


\section{Conclusions}
\label{sec:conclusions}

In this paper, we have proposed an algorithm that performs feature selection for CRFs. The
benefits of working with sparse feature vectors are twofold: obviously, less features need to be
computed and stored; more importantly, sparsity can be used to speed up the forward-backward and
the Viterbi algorithms. Our method combines the $\ell^1$ and $\ell^2$ penalty terms and can thus be
viewed as an extrapolation of the elastic net concept to CRFs. To make the method tractable, we
have develop a sparse version of the forward-backward recursions; we
have also introduced and validated two novel ideas: an approximation of the second order derivative of
the CRF logarithmic loss as well as an efficient parameter blocking scheme. This method has been
tested on artificial and real-world data, using large to very large
feature sets containing more than one hundred million features, and yielding accuracy that is comparable with conventional
training algorithms, and much sparser parameter vectors.

The results obtained in this study open several avenues that we wish to explore in the future. A
first extension of this work is related to finding the optimal weight for the penalization term,
a task that is usually achieved through heuristic search for the value(s) that will deliver the
best performance on a development set. Based on our experiments, this search can be performed
efficiently using pseudo regularization-path techniques, which amount here to start the training
with a very constrained model, and to progressively reduce the weight of the $\ell^1$ term so as to
increase the number of active features. This can be performed effectively at very little cost by
restarting the blockwise optimization from the parameter values obtained with the previous weights
setting (so called ``warm-starts''), thereby greatly reducing the number of iterations needed to
reach convergence.
A second interesting perspective, aiming at improving the training speed, is based on the
observation that after a dozen iterations or so, the number of active features is decreasing
steadily. This suggests that those features that are inactive at that stage will remain inactive
till the convergence of the procedure. Hence, in some situations, limiting the updates to the
features that are currently active can be an efficient way of improving the training speed.
Finally, the sparse forward-backward implementation appears to be most attractive when the number
of labels is very large. Hence, extensions of this idea to cases where the features include, for
instance, conjunctions of tests that operate on more than two successive labels are certainly
feasible. The perspective here consists in taking profit of the sparsity to allow for inclusion of
longer range label dependencies in CRFs.

\bibliographystyle{abbrv}
\bibliography{SLCY09}

\end{document}